# RUSSE'2018: A SHARED TASK ON WORD SENSE INDUCTION FOR THE RUSSIAN LANGUAGE[1]

**Panchenko A.** (panchenko@informatik.uni-hamburg.de)
University of Hamburg, Hamburg, Germany

**Lopukhina A.** (alopukhina@hse.ru)
National Research University Higher School of Economics, Moscow, Russia
The Vinogradov Institute of the Russian Language, Russian Academy of Sciences, Moscow, Russia

**Ustalov D.** (dmitry@informatik.uni-mannheim.de)
University of Mannheim, Mannheim, Germany
Krasovskii Institute of Mathematics and Mechanics, Yekaterinburg, Russia

**Lopukhin K.** (kostia.lopuhin@gmail.com)
Scrapinghub, Moscow, Russia

**Arefyev N.** (nick.arefyev@gmail.com)
Lomonosov Moscow State University, Moscow, Russia
Samsung Moscow Research Center, Moscow, Russia

**Leontyev A.** (aleksey_l@abbyy.com)
ABBYY, Moscow, Russia

**Loukachevitch N.** (louk_nat@mail.ru)
Lomonosov Moscow State University, Moscow, Russia

The paper describes the results of the first shared task on word sense induction (WSI) for the Russian language. While similar shared tasks were conducted in the past for some Romance and Germanic languages, we explore the performance of sense induction and disambiguation methods for a Slavic language that shares many features with other Slavic languages, such as rich morphology and virtually free word order. The participants were asked to group contexts of a given word in accordance with its senses that were not provided beforehand. For instance, given a word "bank" and a set of contexts for this word, e.g. "*bank* is a financial institution that accepts deposits" and "river *bank* is a slope beside a body of water", a participant was asked to cluster such contexts in the *unknown in advance* number of clusters corresponding to, in this case, the "company" and the "area" senses of the word "bank". For the purpose of this evaluation campaign, we developed three new evaluation datasets based on sense inventories that have different sense granularity. The contexts in these datasets were sampled from texts of Wikipedia, the academic corpus of Russian, and an explanatory dictionary of Russian. Overall, 18 teams participated in the competition submitting 383 models. Multiple teams managed to substantially outperform competitive state-of-the-art baselines from the previous years based on sense embeddings.

---

[1] The authors listed in random order as they contributed equally to this study.

# RUSSE'2018: ДОРОЖКА ПО ИЗВЛЕЧЕНИЮ ЗНАЧЕНИЙ СЛОВ ИЗ ТЕКСТОВ РУССКОГО ЯЗЫКА[2]

**Панченко А.** (panchenko@informatik.uni-hamburg.de)
Гамбургский университет, Гамбург, Германия

**Лопухина А.** (alopukhina@hse.ru)
НИУ «Высшая школа экономики», Москва, Россия
Институт русского языка им. В. В. Виноградова РАН, Москва, Россия

**Усталов Д.** (dmitry@informatik.uni-mannheim.de)
Университет Мангейма, Мангейм, Германия
Институт математики и механики им. Н. Н. Красовского
УрО РАН, Екатеринбург, Россия

**Лопухин К.** (kostia.lopuhin@gmail.com)
Scrapinghub, Москва, Россия

**Арефьев Н.** (nick.arefyev@gmail.com)
Московский Государственный Университет им. М. В. Ломоносова, Москва, Россия
Московский Исследовательский Центр Самсунг, Москва, Россия

**Леонтьев А.** (aleksey_l@abbyy.com)
ABBYY, Москва, Россия

**Лукашевич Н.** (louk_nat@mail.ru)
Московский государственный университет им. М. В. Ломоносова, Москва, Россия

В статье описываются результаты первого соревнования по автоматическому извлечению значений слов из неразмеченного корпуса текстов для русского языка. Подобные соревнования проводились для некоторых романских и германских языков; мы исследуем методы извлечения значений и разрешения многозначности на материале одного из славянских языков, обладающих богатой морфологией и достаточно свободным порядком слов. Участникам соревнования было предложено сгруппировать контексты слова в соответствии с его значениями, причем сами значения необходимо было автоматически извлечь из корпуса текстов. Например, для неоднозначного слова «замок» нужно было выделить неизвестное заранее число кластеров, соответствующее его значениям, и классифицировать контексты этого слова так, чтобы каждый контекст попал в тот или иной кластер, соответствующий значению слова — «сооружение» и «устройство, препятствующее доступу куда-либо» для контекстов слова «замок». Для оценки качества работы методов мы подготовили три набора данных, различающихся, во-первых, гранулярностью значений и, во-вторых, источниками контекстов (статьи русскоязычной Википедии, материалы Национального корпуса русского языка и толкового словаря). В соревновании приняли участие 18 команд, приславших 383 моделей. Качество результата, полученного представленными моделями, превосходят эталонные методы, основанные на векторах смыслов.

**Ключевые слова:** лексическая семантика, извлечение смыслов, разрешение лексической многозначности, полисемия, омонимия

---

[2] Все авторы внесли равный вклад в работу; порядок авторов выбран случайным образом.

## 1. Introduction

RUSSE[3] is a series of workshops on evaluation of semantic models for the Russian language. The first workshop on semantic relatedness and similarity was held in 2015 in conjunction with the Dialogue conference [Panchenko et al., 2016][4]. The second event, described in this paper, is dedicated to Word Sense Induction (WSI). Word sense induction is the process of automatic identification of word senses in raw corpora. While evaluation of various sense induction and disambiguation approaches was performed in the past for the Western European languages, e.g., English, French, and German, no systematic evaluation of WSI techniques for Slavic languages are available at the moment. This shared task makes a first step towards bridging this gap by setting up an evaluation campaign for one Slavic language. The goal of this campaign is to compare sense induction systems for the Russian language. Many Slavic languages[5] still do not have broad coverage lexical resources available in English, such as WordNet, which provide a comprehensive inventory of word senses. Therefore, word sense induction methods investigated in this shared task can be of great value to enable semantic processing of under-resourced Slavic languages and domains.

The contribution of our work is two-fold: First, we present a first shared task on word sense induction for a Slavic language. Second, we present three novel sense annotated datasets with about 17 thousands sense-annotated contexts from three sense inventories.

This paper is organised as follows: In Section 2, we describe previous shared tasks covering other languages. In Section 3, we outline the proposed evaluation methodology. Section 4 describes three evaluation datasets. Section 5 presents top scored systems participated in the task. Finally, Section 6 summarizes key results of the shared task.

## 2. Related Work

In this section, we start with an overview shared tasks on word sense induction implying no sense inventory is provided. All prior shared task on this topic were conducted for the English language during the SemEval competitions. Next, we briefly overview previous approaches for word sense disambiguation and induction for the Russian language.

### 2.1. Shared Tasks on Word Sense Induction

In 2007, SemEval participants were provided with 100 target words (65 verbs and 35 nouns), each target word having a set of contexts where the word appears [Agirre and Soroa, 2007]. A part of these contexts was given as a train set, the rest served as a test set. Average number of senses in the dataset was 3.68 per word. Two evaluation scenarios were proposed. The first scenario is the evaluation of the induced senses as clusters of examples. The obtained clusters were compared to the sets of examples labeled with the given gold standard word senses (classes), and evaluated using the clustering measure called FScore. FScore is calculated as the average of the best F-measure values for each cluster relative to the gold standard classes. The second scenario is the mapping of the induced senses to the gold standard senses and using this mapping to label the test corpus with gold standard labels. The results are evaluated with the precision and recall measures for supervised word sense disambiguation systems. It was found that the FScore measure penalized systems with a high number of clusters, and favored those that induced less senses. Supervised evaluation seemed to be more neutral regarding the number of clusters, as the ranking of systems according to this measure include diverse cluster average. So the ranking of the systems varies according to the used evaluation method.

---

[3]   https://russe.nlpub.org

[4]   http://www.dialog-21.ru/en

[5]   http://sigslav.cs.helsinki.fi

In 2010, a similar evaluation was devoted to word sense induction for 100 words [Manandhar et al., 2010]: 50 nouns and 50 verbs. For each target word, participants were provided with a training set in order to learn the senses of that word. Then, participating systems disambiguate unseen instances (contexts) of the same words using the learnt senses. The organizers used two other measures of evaluation in comparison to the 2007 task: paired F-score calculated as F-measure of example pairs included or not-included in the induced clusters; and V-measure that assessed the quality of a clustering solution by explicitly measuring its homogeneity and its completeness according to gold standard classes. It was found that V-measure tended to favor systems producing a higher number of clusters. The organizers concluded that the current state-of-the-art lacks unbiased measures that objectively evaluate clustering.

In 2013, the evaluation was focused on the multi-sense labeling task [Jurgens and Klapaftis, 2013]. In this setup, participating systems provide a context with one or more sense labels weighted by the degree of applicability, which implies the use of fuzzy clustering methods. Measuring the quality of clustering requires handling overlapping clusters, for which two new evaluation measures have been proposed: fuzzy B-Cubed and fuzzy normalized mutual information.

**2.2. Word Sense Disambiguation and Induction for Russian**

For Russian, Loukachevitch and [Chuiko 2007] studied the all-word disambiguation task on the basis of the RuThes thesaurus. They experimented with various parameters (types of the thesaurus paths, window size, etc). [Kobritsov et al. 2005] developed disambiguation filters to provide semantic annotation for the Russian National Corpus[6]. The semantic annotation was based on the taxonomy of lexical and semantic facets. In [Lyashevskaya and Mitrofanova, 2009], statistical word sense disambiguation methods for several Russian nouns were described.

For word sense disambiguation, word sense frequency information is very important. Loukachevitch and [Chetviorkin 2015] studied the approach of determining the most frequent sense of ambiguous words using unambiguous related words and phrases described in the RuThes thesaurus. [Lopukhina et al., 2018] estimated sense frequency distributions for noun taken from the to Active Dictionary of Russian.

---

[6] http://ruscorpora.ru/en

Concerning word sense induction task for Russian, [Lopukhin et al. 2017] evaluated four methods: Adaptive Skip-gram, Latent Dirichlet Allocation, clustering of contexts, and clustering of synonyms. [Ustalov et al. 2017] proposed a fuzzy graph clustering algorithm Watset designed for unsupervised acquisition of word senses and grouping them into sets of synonyms (synsets) using semi-structured dictionaries, such as Wiktionary and synonymy dictionaries.

## 3. Shared Task Description

This shared task is structurally similar to prior WSI tasks for the English language, such as SemEval 2007 WSI [Agirre and Soroa, 2007][7] and SemEval 2010 WSI&D [Manandhar et al, 2010][8] tasks. Namely, we rely on the "lexical sample" setting, where participants are provided with a set of polysemous words, each word is provided with a set text fragments called *contexts* representing examples of the word usage in various senses.

For instance, the contexts for the word "bank" can be "In geography, the word *bank* generally refers to the land alongside a body of water" and "The *bank* offers financial products and services for corporate and institutional clients". For each context, a participant specifies the sense of the target word. Note that we do not provide any sense inventory: the participants can assign sense identifiers of their choice to a context, e.g., "bank#1" or "bank (area)". The only requirement is that the contexts with the different senses of the target word should be assigned with the different identifiers, while the contexts representing the same senses should be assigned with the same identifier. In our study, we use the word "context" as the synonym of the word "instance" used in SemEval [Agirre and Soroa, 2007]; [Manandhar et al, 2010]. Detailed instructions for participant were provide on the shared task[9] website and in the GitHub repository.[10]

### 3.1. Tracks

We distinguish two tracks in RUSSE'2018. In the *knowledge-free* track, the participants induce a sense inventory from any text corpus of their choice and use this inventory for assigning sense identifiers to the contexts. In the *knowledge-rich* track, the participants use an existing sense inventory, i.e., a dictionary, to disambiguate the target words. The use of the gold standard inventories are prohibited in both tracks.

---

[7]  http://semeval2.fbk.eu/semeval2.php?location=tasks&taskid=2

[8]  https://www.cs.york.ac.uk/semeval2010_WSI

[9]  https://russe.nlpub.org/2018/wsi/

[10]  https://github.com/nlpub/russe-wsi-kit

The advantage of our setting is that virtually any existing word sense induction approach can be used within the framework of our shared task, starting from unsupervised sense embeddings to the graph-based methods that rely on lexical knowledge bases, such as WordNet.

### 3.2. Evaluation Datasets

We provide three labeled datasets with contexts sampled from different text sources, which are based on different sense inventories. Each of the dataset was split into the train and test sets. Both sets use the same corpora and annotation principles, but the target words are different. The train set was given to the participants for tuning their models before the competition starts. The test set was made available without labels at the end of the competition. We provide an extensive description of the datasets in Section 4.

### 3.3. Quality Measure

Similarly to SemEval 2010 Task 14 and SemEval 2013 Task 13 on word sense induction and disambiguation, we use a gold standard, in which each polysemous target word is provided with a set of contexts. Each context is manually annotated with a sense identifier as according to the predefined sense inventory. A participating system assigns the sense identifiers from the chosen sense inventory to these ambiguous contexts, which can be seen as clustering of contexts. Thus, to evaluate a system, the labeling of contexts provided by the system is compared to the gold standard labeling, although the sense inventories are different.

Clustering-based measures have an important constraint: they provide contradictory rankings. For instance, none of the five evaluation measures in the SemEval 2013 shared task agree to each other, preferring larger or smaller clusters, see [Jurgens and Klapaftis 2013]. In our shared task, we wanted to avoid having multiple evaluation measures that may provide conflicting results. Moreover, we wanted to have a measure which is equal to zero in the cases of trivial clustering, i.e., random clustering, separate cluster for each context, single cluster for all contexts. We selected a measure that fits all these demands, namely the Adjusted Rand Index (ARI) by [Hubert and Arabie, 1985]. We adopted ARI implementation from the scikit-learn library[11]. The measure was also used before for evaluation of word sense induction in SemEval 2013 Task 11 [Navigli and Vannella, 2013] and in [Bartunov et al., 2016].

### 3.4. Baseline Systems

We provided a state-of-the-art baseline based on unsupervised word sense embeddings called AdaGram [Bartunov et al., 2016], which is a multi-prototype Bayesian extension of the Skip-gram model [Mikolov et al., 2013]. We rely on a model by [Lopukhin et al. 2017] trained on the 2B tokens-large lemmatized corpus combining the ruWac Internet corpus [Sharoff, 2006], the Russian online library lib.ru, and the Russian Wikipedia. The baseline was obtained using the following hyperparameters: the maximum number of senses of 10, the sense granularity of 0.1, the vector dimension of 300, and the context window of 5. No additional tuning of baseline method on the train data was performed; its performance could be further improved by adjusting the of number of senses for each dataset, merging of similar senses and weighting the contexts. In addition to AdaGram, we provided trivial baselines based on random assignment of word senses, putting each context into a singleton cluster, and putting all the contexts of a word into the same cluster.

---

[11] http://scikit-learn.org/stable/modules/generated/sklearn.metrics.adjusted_rand_score.html

## 4. Evaluation Datasets

We prepared three new gold standard datasets for RUSSE'2018. These datasets are complementary in terms of the average number of senses per word (granularity) of their sense inventories and in terms of the text corpora from which the contexts were sampled. Each of these datasets is named by *corpus-inventory* principle. We have also provided the participants with three published datasets from [Lopukhin and Lopukhina, 2016] as a source of additional training data. Statistics for all the datasets used in the shared task are presented in Table 1.

**Table 1.** The datasets used in the shared task. The "main" datasets were used to test the runs of the participants, and the "additional" datasets were provided as a source of extra training data

| Dataset | Type | Inventory | Corpus | Split | # of words | # of senses | Avg. # of senses | # of contexts |
|---|---|---|---|---|---|---|---|---|
| wiki-wiki | main | Wikipedia | Wikipedia | train | 4 | 8 | 2.0 | 439 |
| wiki-wiki | main | Wikipedia | Wikipedia | test | 5 | 12 | 2.4 | 539 |
| bts-rnc | main | BTS | RNC | train | 30 | 96 | 3.2 | 3491 |
| bts-rnc | main | BTS | RNC | test | 51 | 153 | 3.0 | 6556 |
| active-dict | main | Active Dict. | Active Dict. | train | 85 | 312 | 3.7 | 2073 |
| active-dict | main | Active Dict. | Active Dict. | test | 168 | 555 | 3.3 | 3729 |
| active-rnc | additional | Active Dict. | RNC | train | 20 | 71 | 3.6 | 1829 |
| active-rutenten | additional | Active Dict. | ruTenTen[12] | train | 21 | 71 | 3.4 | 3671 |
| bts-rutenten | additional | BTS | ruTenTen | train | 11 | 25 | 2.3 | 956 |

---

[12] The ruTenTen11 is a large web-based corpus of Russian consisting of 18 billion tokens, which is available thought the Sketch Engine system [Kilgarriff et al., 2004].

### 4.1. *wiki-wiki:* A Dataset Based on Wikipedia

This sense inventory was built from scratch using words from homonymous word forms dictionary[13] and their senses occurred in the Russian Wikipedia article titles. The contexts have been extracted from the Russian Wikipedia. We assumed that given a Wikipedia article containing an ambiguous word in its title, all the occurrences of this word in this article will share the same sense. Hence, we manually assigned sense identifiers to the titles and extracted contexts of these senses from the full texts of the articles automatically. The datasets contains 9 nouns with 20 homonymous senses.

To construct the dataset, list of the Russian Wikipedia articles which titles contain homonyms from the dictionary has been created. These homonyms which do not occur in the article titles or occurred less than 40 times in the corresponding articles were excluded. The titles for each of the remaining words were grouped manually as according to the homonym sense. Each sense was described using related words (synonyms, antonyms, associations etc.) from the Russian Wiktionary. This resulted in the sense inventory an excerpt of which is presented in Table 2.

**Table 2.** An excerpt from the sense inventory of wiki-wiki dataset: the word "белка"

| word | articles | sense |
| --- | --- | --- |
| белка | кавказская белка; обыкновенная белка; японская белка; капская земляная белка; аризонская белка; ... | рыжая, шустрая, дерево, вскарабкаться, спрыгнуть |
| белка | домен белка; биосинтез белка; фолдинг белка; институт белка ран; сигнальная функция белка; ... | желток, пища, углевод, рацион, жир |
| белка | белка и стрелка; белка и стрелка (мюзикл); белка и стрелка. лунные приключения; ... | космос, полет, животные, первые, советские |

Then, for each sense of each homonym we parsed full texts of the corresponding articles and extracted each occurence of the homonym with at least 50 words to the left and at least 50 words to the right to form a context. If we found no at least 10 contexts for any sense of a homonym, we excluded it with all its senses from the dataset to keep the dataset balanced. Finally, all the contexts have been verified by the organizers; only 9 out of 15 homonyms were left.

### 4.2. *bts-rnc:* A Dataset Based on the Russian National Corpus

This dataset is based on the sense inventory of the Large Explanatory Dictionary of Russian[14] (*Bolshoj Tolkovyj Slovar'*, BTS; [Kuznetsov, 2014]). The contexts were sampled from the Russian National Corpus (RNC, 230 million tokens in the main corpus)[15]. The train set contains 30 ambiguous words: 9 polysemous words with metaphorical senses and 21 homonymous word[16]. The test set contains 51 ambiguous words: 11 polysemous words with metaphorical senses and 40 homonymous word. We selected these two types of ambiguity—homonymy and metaphorical extension in polysemy—because they were proven to be distinguishable by native speakers in psycholinguistic experiments [Klein and Murphy, 2001]; [Klepousniotou, 2002]; [Klepousniotou et al., 2008]. In this shared task, 29 out of 61 homonyms have only one sense each (e.g. "крона" as a "crown of a tree" and "крона" as in "Norwegian or Danish krone"), the other 32 homonyms were polysemous (e.g. "икра" as "roe / caviar" or "eggplant paste" and "икра" as "calf of a leg"). So we assumed they might be also distinguishable in language models.

---

[13] http://cfrl.ruslang.ru/homoforms/index.htm

[14] http://gramota.ru/slovari/dic

The dataset was manually annotated by four students majoring in linguistics. Then, the experts checked the annotation and fixed the mistakes. To ensure the high quality of the annotated dataset, we invited expert linguists for a systematic check of every annotated context for complex words with a high number of polysemous senses. For simpler words with a small number of homonymous senses, we used microtask-based crowdsourcing. Namely, 20 words and 2547 contexts were checked using crowdsourcing, and 61 words and 7500 contexts were checked by 7 human experts, which are the authors of this paper. Each human expert read all the contexts and fixed wrong sense annotations, or removed contexts which were too ambiguous or simply irrelevant, e.g., in the cases when a real sense mentioned in the context was actually not in the sense inventory. Overall, 2103 out of 12150 contexts were removed, such as an irrelevant context for the word "гвоздика" presented below. This word representing "flower" and "spice" senses are confused in this context with its homograph "гвоздик" (nail):

> … Посмотри, как здорово это будет выглядеть! — Хорошо, а <u>гвоздики</u> для картин вы сами в стенку вбиваете? …
>
> … Look how great it will look! — Well, do you drive <u>nails</u> for the pictures into the wall?

Another example of the filtered sentences, is with an ambiguous context for the word "крыло" (wing) where the described situation is unclear:

> … волны, а чуть противный ветер, и <u>крылья</u> повисли; рядом же мчится, несмотря ни на что, пароход, и человек сидит
>
> … waves, but a slightly nasty wind, and the <u>wings</u> are hanging down; nevertheless, a steamship is racing alongside and a man is sitting

---

[15] http://ruscorpora.ru

[16] In the case of homonymy, a lexical item carries two (or more) distinct unrelated meanings, such as bank as a financial institution and bank as a side of a river; in the case of polysemy senses of a word are related, e.g. blood in "His face was covered in blood" and "They had royal blood in their veins". [Lyons, 1977]

The crowdsourcing annotation was performed on Yandex.Toloka platform[17]. For annotation, we used a subset of words with two or three distinct meanings. In this task, a crowd worker is provided with a set of contexts with a highlighted word to be disambiguated. The worker chooses one of the senses listed below the sentence and submits the answer. The workers demonstrated a high inter-annotator agreement as according to the Krippendorff's α value of 0.825 [Krippendorff, 2013].

### 4.3. *active-dict*: A Dataset Based on a Dictionary

The Active Dictionary of Russian is an explanatory dictionary that has a strong theoretical basis in sense distinction and reflects contemporary language. (*Aktivnyj slovar' russkogo jazyka*; [Apresjan, 2014]; [Apresjan et al., 2017]). The word senses in the Active Dictionary are considered distinct if they have different semantic and syntactic properties, collocational restrictions, synonyms, and antonyms. For each sense, we extracted all examples (short and common usages) and illustrations (longer, full-sentence examples from the Russian National Corpus) that were used as context in this shared task. On average, we extracted 22.9 contexts per word. The train set, having 85 ambiguous words (84 polysemous words and 1 homonym) and 2073 contexts, was extracted from publicly available first and second volumes of the dictionary (letters A–G; [Apresjan, 2014]). The test set, having 168 ambiguous words (167 polysemous words and 1 homonym), and 3729 contexts, was taken from the third volume of the dictionary that became available in March 2018 (letters D–Z; [Apresjan, Galaktionova, Iomdin, 2017]).

To construct the dataset, we extracted examples and illustrations for all polysemous nouns and merged homonymous nouns together. The parser inputs an unstructured representation of the dictionary in a word processor format and outputs a set of labeled contexts.

## 5. Participating Systems

Overall 18 teams participated in the RUSSE'2018 shared task. We provide here self-descriptions of the approaches used by the teams ranked within the top 5 list in each dataset. The descriptions for all the models submitted by the participants for all datasets can be found in the CodaLab platform in the "Public Submissions" section. We denote each team with its CodaLab login, e.g., "jamsic", and also provide a reference to the paper describing the approach, where available. Note that all the participants submitted to the "knowledge-free" track and we received no submissions to the "knowledge-rich" track.

---

[17] https://toloka.yandex.ru

### 5.1. The *wiki-wiki* Dataset based on Wiktionary

17 teams submitted 124 runs for this dataset, with the top teams being:

- **jamsic**. This team used a pre-trained CBOW word embeddings model with 300 dimensions based on the Russian National Corpus by [Kutuzov and Andreev, 2015][18]. The sense clusters were obtained directly from this model by looking at the list of the nearest neighbours. The approach identifies two senses per word. First, the most most similar term to a target word is retrieved. This word represents the first sense. Second, vector representation of this word is subtracted from the vector of the target word and again the most similar term is retrieved. This second term represents the second word sense. Disambiguation of a context is performed via calculation of cosine distance of a context representation (an average of embeddings) with these two prototypes. [Pelevina et al. 2016] proposed another method for induction of senses from word embeddings which used clustering of ego-network of related words. However, this approach does not make use of vector subtraction operation employed by the *jamsic* team.
- **akutuzov** [Kutuzov, 2018]. This team used Affinity Propagation to cluster weighted average of word embeddings for each context. The embedding model was trained on the Russian National Corpus using a newer version of the embeddings as compared to [Chernobay, 2018].
- **ezhick179** [Arefyev et al., 2018]. This team used Affinity Propagation to cluster the non-weighted average of CBOW vectors for contexts trained on a large corpus of Russian books based on the lib.rus.ec collection with the vector dimensions of 200, the context window of 10, in 3 iterations [Arefyev et al., 2015].[19]
- **aby2s**. This team relied on hierarchical clustering of context embeddings based on the Ward clustering with cophenetic distance criterion and a threshold of 2.6. Sentences were represented as normalized sums of fastText [Bojanowski et al., 2016] embeddings pre-trained on a Wikipedia corpus.
- **Pavel** [Arefyev et al., 2018]. This team used agglomerative clustering of the weighted average of Word2Vec vectors for contexts. The words were weighed using the $tfidf^{1.5} \times chisq^{0.5}$ score. The word embeddings were the CBOW vectors for contexts trained on lib.rus.ec with the vector dimensions of 200, the context window of 10, in 3 iterations [Arefyev et al., 2015].

### 5.2. The "bts-rnc" Dataset based on the Russian National Corpus

16 teams submitted 121 runs for this dataset, with the top teams being:

- **jamsic, akutuzov, ezhick179, Pavel** used methods described in Section 5.1.
- **fogside**: Used word embeddings trained on a combination of Wikipedia, Librusec and the training dataset. A neural network with self-attention was used to encode the sentence representations, which were subsequently clustered with the k-means algorithms with k=2.

---

[18] http://rusvectores.org

[19] https://nlpub.ru/RDT

## 5.3. The "active-dict" Dataset based on a Dictionary

18 teams submitted 138 runs for this dataset, with the top teams being:

- **jamsic**. This team used a pre-trained CBOW word embeddings model of 300 dimensions based on the Russian National Corpus by [Kutuzov and Andreev, 2015] as in the previous two datasets. However, in this submission the authors followed the approach to word sense embeddings proposed by [Li and Jurafsky, 2015].
- **akutuzov, ezhick179, Pavel**: These teams used methods described in Section 5.1.

## 6. Results

Tables 2, 3, and 4 present the results of the shared task for the three datasets used for evaluation: *wiki-wiki*, *bts-rnc*, and *active-dict*. Each table lists top 10 best teams with the public and private ARI scores on the test set (see Section 4). We disregarded from the final ranking teams which were created by organizers for testing purposes[20] and teams which did not provide any description of the used approach[21]. The private ARI scores are used for final ranking of the participants, while the public scores were visible to the participants on the leaderboard immediately after submission before the final deadline. Private and public scores were calculated on non-overlapping sets of words, with public words constituting approximately one third of all words in test set of each dataset. Public scores allowed participants to immediately see their position relative to other participants, while using private scores for final evaluation ensured that participants did not pick their submission based on the leaderboard score. A large difference in public and private scores for the *wiki-wiki* dataset is due to the public set consisting of contexts for just two words: this caused large variance between the public and the private parts for this dataset. However, private/public test differences are substantially smaller for other larger datasets.

**Table 3.** Top 10 teams out of 17 on the "wiki-wiki" dataset.
The full table is available at the CodaLab platform:
https://competitions.codalab.org/competitions/17810#results

| Rank | Team | ARI (public) | ARI (private) |
|---|---|---|---|
| 1 | jamsic | 1.0000   (1) | 0.9625   (1) |
| 2 | akutuzov [Kutuzov, 2018] | 0.9823   (2) | 0.7096   (2) |
| 3 | ezhick179 [Arefyev et al., 2018] | 1.0000   (1) | 0.6586   (3) |
| - | *akapustin* | 0.6520   (6) | 0.6459   (4) |
| - | *aby2s* | 1.0000   (1) | 0.5889   (5) |
| - | *bokan* | 0.7587   (5) | 0.5530   (6) |
| * | **AdaGram [Bartunov et al, 2016]** | **0.6278   (7)** | **0.5275   (7)** |
| 4 | Pavel [Arefyev et al., 2018] | 0.9649   (3) | 0.4827   (8) |
| 5 | eugenys | 0.0115 (12) | 0.4377   (9) |
| 6 | mikhal | 1.0000   (1) | 0.4109 (10) |
| 7 | fogside | 0.6520   (6) | 0.3958 (11) |

---

[20] Team names: russewsi, lopuhin, panchenko, dustalov.

[21] Team names: joystick, Timon, thebestdeeplearningspecialist, bokan, akapustin, ostruyanskiy.

**Table 4.** Top 10 teams out of 16 on the "bts-rnc" dataset.
The full table is available at the CodaLab platform:
https://competitions.codalab.org/competitions/17809#results

| Rank | Team | ARI (public) | ARI (private) |
|---|---|---|---|
| 1 | jamsic | 0.3508  (1) | 0.3384  (1) |
| 2 | Pavel [Arefyev et al., 2018] | 0.2812  (2) | 0.2818  (2) |
| - | *joystick* | 0.2477  (5) | 0.2579  (3) |
| - | *Timon* | 0.2360  (7) | 0.2434  (4) |
| 3 | akutuzov [Kutuzov, 2018] | 0.2448  (6) | 0.2415  (5) |
| 4 | ezhick179 [Arefyev et al., 2018] | 0.2599  (4) | 0.2284  (6) |
| - | *thebestdeeplearningspecialist* | 0.2178  (8) | 0.2227  (7) |
| 5 | fogside | 0.1661 (10) | 0.2154  (8) |
| * | **AdaGram [Bartunov et al., 2016]** | 0.2624  (3) | 0.2132  (9) |
| - | *aby2s* | 0.1722  (9) | 0.2102 (10) |
| 6 | bokan | 0.1363 (11) | 0.1515 (11) |

**Table 5.** Top 10 teams out of 16 on the "active-dict" dataset.
The full table is available at the CodaLab platform:
https://competitions.codalab.org/competitions/17806#results

| Rank | Team | ARI (public) | ARI (private) |
|---|---|---|---|
| 1 | jamsic | 0.2643  (1) | 0.2477  (1) |
| 2 | Pavel [Arefyev et al., 2018] | 0.2361  (4) | 0.2270  (2) |
| - | *Timon* | 0.2324  (5) | 0.2222  (3) |
| - | *thebestdeeplearningspecialist* | 0.2297  (6) | 0.2194  (4) |
| 3 | akutuzov [Kutuzov, 2018] | 0.2396  (3) | 0.2144  (5) |
| - | *aby2s* | 0.2465  (2) | 0.1985  (6) |
| - | *joystick* | 0.1890  (8) | 0.1939  (7) |
| 4 | ezhick179 [Arefyev et al., 2018] | 0.1899  (7) | 0.1839  (8) |
| * | **AdaGram [Bartunov et al., 2016]** | 0.1764  (9) | 0.1538  (9) |
| - | *ostruyanskiy* | 0.1515 (10) | 0.1403 (10) |
| - | *akapustin* | 0.1337 (11) | 0.1183 (11) |

Several observations can be made on the basis of the results presented in the Tables 3–5. First, the method of **jamsic**, based on extraction of sense inventory directly from word sense embeddings showed good results on two datasets were it was applied substantially outperformed all other methods (see Section 5 for a detailed description of the methods). A particularly large advantage of this method over other method, which relied on some kind of sentence clustering, is observed for the coarse-grained wiki-wiki dataset, because it contains only homonymous senses, which can be easily extracted with such an approach. On the *active-dict* dataset this participant also outperformed other teams, but in this case using the approach of [Li and Jurafsky, 2015] to the construction of word sense embeddings.

Second, other approaches showing good results ranking in top 2–5 were the methods based on direct clustering of textual contexts represented with the features based on word embeddings pre-trained on large corpora, such as the Russian National Corpus or a collection of books from the lib.rus.ec library. In particular, successful methods relied on the Affinity Propagation clustering approach, but also some other methods, such as Agglomerative and Ward clustering algorithms. The **fosgide** team used word embeddings and the *k*-means clustering algorithm. Namely, for each context, a context vector is built as a non-weighted average of the fastText vectors for the words in the context. The context vectors for each target word are decomposed into a linear combination of learnt basis vectors. Then, weight vectors of this decomposition are clustered using k-means clustering algorithm.

The Affinity Propagation method is well-suited in the case of word sense induction task as it defines the number of parameters automatically, in contrast to, e.g., Agglomerative Clustering, which produces a lot of senses in the case of this task, as the number of sense per word is usually distributed according to a power law. Nevertheless, the *Pavel* [Arefyev et al., 2018] team managed to obtain two second-best results on two datasets using Agglomerative Clustering with a fixed number of clusters (different in the case of each dataset, learned from the train data). It was shown that a carefully selected weighting schema for words can provide an edge with respect to a un-weighted average of word embeddings. Besides, on *wiki-wiki* and *rnc-bts* datasets, *jamsic* team provided good results with the method which also yields two senses per word for all words. In case of the first dataset, this could be explained by the fact that the average polysemy of this dataset is 2. In the case of the second dataset with more senses, the good performance could be explained by a skewed distribution of senses across the sentences: the majority of the contexts belong to two senses (which is not the case for the sense-balanced *active-dict* dataset).

Third, multiple teams managed to outperform a competitive baseline provided by the organizers based on the AdaGram [Bartunov et al., 2016] word sense embeddings. There is a substantial difference in ARI score between different datasets. The scores for *wiki-wiki* dataset are much higher due to a low number of senses per word and extremely clear separation between senses. Among two other datasets based on dictionary senses, scores for *bts-rnc* are higher than for *active-dict* due to *active-dict* using a more granular sense inventory and having a much smaller number of contexts per sense (just 6.8 instead of 42 for *bts-rnc*), see Table 1 and analysis in [Lopukhin et al., 2017]. Another difference is that for *bts-rnc* contexts were randomly sampled from corpus, while contexts for *active-dict* were selected by the authors of the Active Dictionary of Russian, with both full sentences from the corpus and short usage examples — it remains unclear how this difference contributed to the difference in scores. Still, ARI scores for all datasets are higher than what was reported in [Bartunov et al. 2016] for SemEval-2007 and SemEval-2010 datasets for word sense induction.

Finally, one can observe a large difference in absolute scores for the coarse-grained *wiki-wiki* dataset and the two datasets based on fine-grained word sense inventories coming from dictionaries (*active-dict* and *bts-rnc*). Discriminating between a large number of related polysemous senses is naturally a more challenging task, which requires more sophisticated representations and methods. We hope that the setup of our shared task will pave the way towards developing methods which are able also to excel on these more challenging datasets.

## 7. Conclusion

In this paper, we presented the results of the first shared task on word sense induction for a Slavic language. For this shared task, three new large-scale datasets for word sense induction and disambiguation for the Russian language have been created and published. The shared task attracted 18 participating teams, which submitted overall 383 model runs on these three datasets. A substantial amount of the participant were able to outperform a competitive state-of-the-art baseline approach put in place by the organizers based on the AdaGram word sense embeddings method [Bartunov et al., 2016]. This shared task is the first systematic attempt to evaluate word sense induction and disambiguation systems for the Russian language. We hope that the produced resources and datasets will foster further research and help development of new generation of the sense representation methods.

### Acknowledgements

This research was supported by Deutsche Forschungsgemeinschaft (DFG) under the projects JOIN-T and ACQuA and by the RFBR under the project no. 16-37-00354 мол_а. The work of Konstantin Lopukhin and Anastasiya Lopukhina was supported by a grant of the Russian Science Foundation, Project 16-18-02054. We are grateful to the authors of the Active Dictionary of Russian who kindly allowed us to use the dictionary in this shared task. We are grateful to Ted Pedersen for a helpful discussion about the evaluation settings. Finally, we thank our supporters and information sponsors, who helped to spread the word about the task: ABBYY, The Special Interest Group on Slavic Natural Language Processing (ACL SIGSLAV), Moscow Polytechnic University, Mathlingvo, and NLPub.